\documentclass[conference]{IEEEtran}
\IEEEoverridecommandlockouts
\usepackage{cite}
\usepackage{amsmath,amssymb,amsfonts}
\usepackage{graphicx}
\usepackage{textcomp}
\usepackage{xcolor}
\usepackage{booktabs}
\usepackage{graphicx}
\usepackage{booktabs}
\usepackage{amssymb}
\usepackage{algorithm}
\usepackage{algpseudocode}
\usepackage{wrapfig}
\usepackage[hidelinks]{hyperref}
\def\BibTeX{{\rm B\kern-.05em{\sc i\kern-.025em b}\kern-.08em
    T\kern-.1667em\lower.7ex\hbox{E}\kern-.125emX}}
\begin{document}

\title{Cross-Modality Controlled Molecule Generation with Diffusion Language Model}

\author{
\IEEEauthorblockN{
Yunzhe Zhang,
Yifei Wang,
Andrew Nguyen,
and Pengyu Hong
}
\IEEEauthorblockA{
\textit{Department of Computer Science} \\
\textit{Brandeis University} \\
Waltham, MA, USA \\
\{yunzhezhang, yifeiwang, andrewng, hongpeng\}@brandeis.edu
}
}

\maketitle

\begin{abstract}
The increasing variety of molecular data creates a need for generative models that can flexibly incorporate heterogeneous constraints across modalities. However, existing SMILES-based diffusion models are typically designed for a fixed conditioning modality, and introducing new constraints often requires retraining the model. To address this limitation, we propose Cross-Modality Controlled Molecule Generation with Diffusion Language Model (CMCM-DLM), a modular framework that extends a pre-trained diffusion model to support heterogeneous molecular constraints without retraining the backbone. We demonstrate CMCM-DLM using two complementary modalities: molecular structure and chemical properties. Specifically, a Structure Control Module (SCM) guides early diffusion steps to establish the molecular scaffold, while a Property Control Module (PCM) subsequently steers generation toward target chemical properties. This staged design enables flexible integration of different molecular constraints within a unified generative framework. Experiments on multiple datasets demonstrate effective cross-modal controllability and strong adaptability, highlighting the potential of CMCM-DLM for heterogeneous molecular data modeling and data-driven drug discovery.
\end{abstract}

\begin{figure*}[t] 
    \centering
    \includegraphics[width=\textwidth]{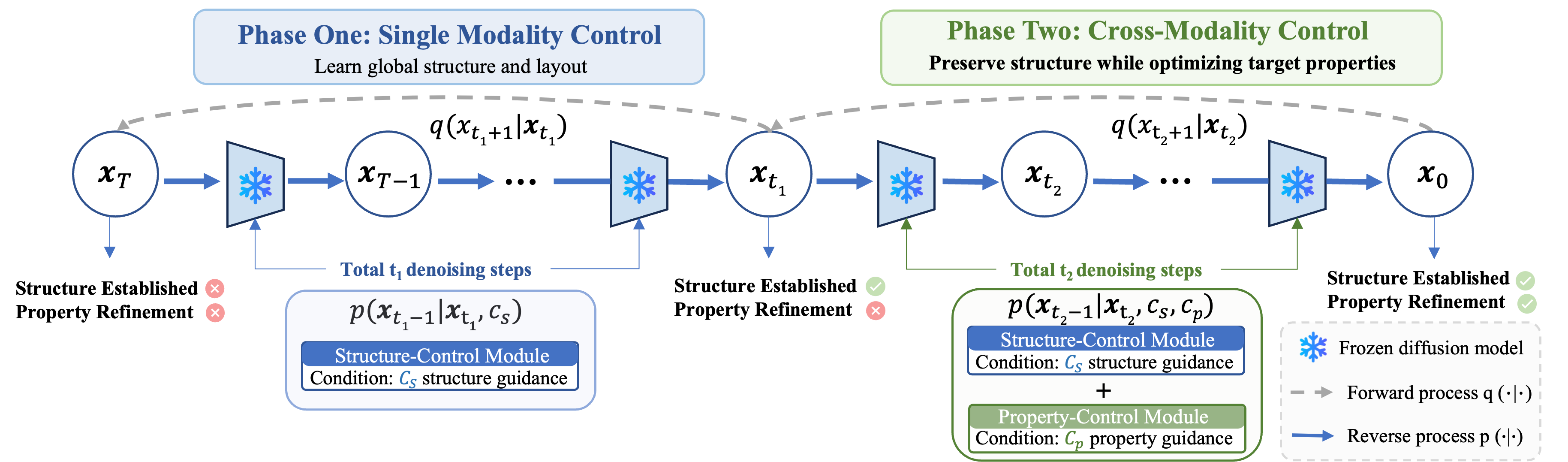} 
    \caption{Overview of CMCM-DLM for adaptive cross-modality molecular generation. Starting from a pre-trained diffusion language model, Phase One introduces a structural constraint $c_s$ to establish the molecular scaffold, while Phase Two incorporates a property constraint $c_p$ to guide generation toward chemical properties. The modular controls are applied during inference without retraining the diffusion backbone.}
    \label{fig:example_image}
    \vspace{-2mm}
\end{figure*}

\section{Introduction}

Modern drug discovery increasingly relies on heterogeneous molecular data, including molecular structures, textual descriptions, and physicochemical properties. Such data provide complementary information for molecular design, but also require generative models to flexibly incorporate constraints across different modalities. This is particularly important in drug discovery, where candidate molecules must often satisfy structural requirements together with multiple chemical properties such as drug-likeness, lipophilicity, and synthetic accessibility. Meanwhile, conventional drug discovery remains labor-intensive and time-consuming \cite{hughes2011principles}, motivating data-driven approaches that can more efficiently explore the vast chemical space \cite{gomez-bombarelli2018automatic}.

SMILES representations \cite{weininger1988smiles,weininger1989smiles} provide a compact sequence-based encoding of molecular structures and enable molecular generation to be formulated as a language modeling problem. Recent diffusion models \cite{ho2020denoising,song2021denoising} have demonstrated strong controllability through iterative generation, and diffusion language models have further been applied to controllable sequence and molecular generation \cite{li2022diffusion,gong2024text}. Recent studies have also explored multimodal molecular representations and multi-property conditional generation \cite{liu2023moleculestm,chang2024molecularfoundation,liu2024graphdit}, highlighting the importance of incorporating heterogeneous molecular information into generative models.

However, existing conditional molecular diffusion models typically define their conditioning space during training, such as text-based or property-based controls \cite{gong2024text,chang2025ldmol,liu2024graphdit}. Extending these models to new constraints can therefore require additional training or adaptation, particularly when new properties or combinations of properties are introduced \cite{han2024trainingfree}. Moreover, structural and property constraints are inherently heterogeneous, representing scaffold requirements and numerical molecular objectives, respectively. This motivates a key question: can a frozen pre-trained molecular diffusion model flexibly incorporate new cross-modal controls without repeatedly retraining the generative backbone?

To address this challenge, we propose \textbf{Cross-Modality Controlled Molecule Generation with Diffusion Language Model (CMCM-DLM)}, a modular framework for integrating heterogeneous molecular constraints into a frozen pre-trained diffusion language model. CMCM-DLM contains two lightweight modules: the Structure Control Module (SCM) and the Property Control Module (PCM). In Phase I, SCM guides early diffusion steps to establish the target molecular scaffold. In Phase II, PCM further steers generation toward desired chemical properties while preserving the structural constraint. This staged design enables flexible integration of different molecular data modalities without retraining the pretrained generative backbone.

As illustrated in Fig.~\ref{fig:example_image}, CMCM-DLM adopts a two-phase control strategy. In Phase I, the SCM guides the frozen diffusion backbone to establish the target molecular structure. In Phase II, the SCM and PCM are jointly applied to preserve the established structure while refining the desired molecular properties. This design provides four practical advantages:
\begin{itemize}
\item \textbf{Plug-and-Play.} SCM and PCM are attached to the frozen diffusion backbone as lightweight control modules, allowing new constraints to be introduced without retraining the backbone.
\item \textbf{Flexible.} The framework supports heterogeneous control signals, including diverse molecular scaffolds and properties such as QED, PLogP, and SAS.
\item \textbf{Composable.} Structural and property controls are integrated in a staged manner, enabling multiple constraints to be combined while preserving the established molecular structure.
\item \textbf{Lightweight.} Only the control modules require training: SCM requires less than one-fifth of the backbone pre-training time, while PCM requires no more than half.
\end{itemize}

Our main contributions are threefold. First, we propose CMCM-DLM, a modular framework that extends a pre-trained diffusion language model to support heterogeneous molecular data and cross-modality constraints without retraining the pretrained generative backbone. Second, we develop a two-phase control strategy that decouples structural and property guidance, allowing different controls to be independently introduced and flexibly composed. Finally, experiments on three benchmark datasets demonstrate effective structural and property control while maintaining molecular validity, novelty, and diversity with low adaptation cost.

\section{Related Works}

\subsection{SMILES-based Molecule Generation}

Molecules can be represented as SMILES, molecular graphs, or 3D geometries~\cite{guo2022graph,hoogeboom2022equivariant}. Among them, SMILES provides a compact sequential representation that naturally supports language-based generative modeling. Early SMILES generation methods mainly relied on RNNs~\cite{segler2017generating} and VAEs~\cite{kusner2017grammar}, while autoregressive models such as MolGPT~\cite{bagal2022molgpt} and ChemGPT~\cite{frey2022neural} leverage Transformer architectures for molecular generation. Diffusion language models provide an alternative to autoregressive generation. TGM-DLM~\cite{gong2024text} applies diffusion language modeling to text-guided SMILES generation, while LDMol~\cite{chang2025ldmol} introduces a structurally informative latent space for text-conditioned molecule generation. In contrast, our work focuses on extending a pre-trained SMILES diffusion model to integrate heterogeneous molecular information through modular controls, addressing the growing diversity of data and constraints in molecular design.

\subsection{Controllable Molecule Generation}

Controllable molecule generation aims to integrate diverse molecular information, including structural, property, and semantic constraints, into the generation process. Scaffold-based methods enforce structural requirements~\cite{lim2020scaffold}, while models such as MolGPT support property-conditioned generation~\cite{bagal2022molgpt}. Recent studies have further explored heterogeneous molecular information: MoleculeSTM~\cite{liu2023moleculestm} aligns molecular structures with textual descriptions, SPMM~\cite{chang2024molecularfoundation} jointly models molecular structures and biochemical properties, and Graph DiT~\cite{liu2024graphdit} enables multi-property conditional generation. However, these approaches generally define their supported conditioning space during training, limiting adaptability when new data modalities or design constraints emerge. Our work instead enables modular cross-modality control over a frozen pre-trained diffusion language model, allowing heterogeneous constraints to be independently introduced and composed without retraining the pretrained generative backbone.

\begin{figure*}[t] 
    \centering
    \includegraphics[width=\textwidth]{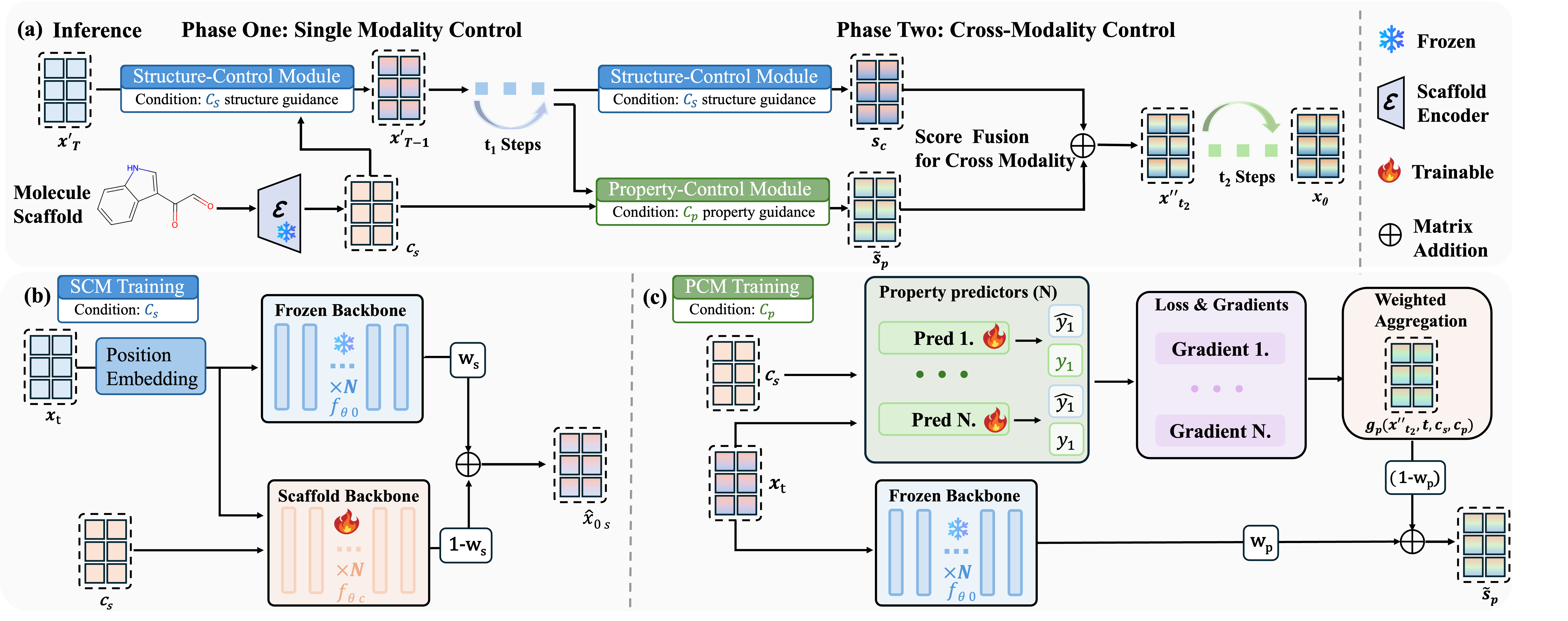} 
    \vspace{-6mm}
    \caption{The CMCM-DLM consists of two phases illustrated in the upper part of the figure. Phase One performs single-modality control using the Structure Control Module (SCM). Phase Two enables cross-modality control by incorporating both the SCM and the Property Control Module (PCM), with their internal architectures shown in the lower part of the figure when the corresponding parameters are provided.}
    \vspace{-6mm}
    \label{fig:framework}
\end{figure*}

\section{Preliminaries and Task Formulation}
\label{sec:preliminary}

This section introduces the key components of the molecular generation pipeline using a diffusion language model (DLM) for SMILES generation, adapted from \cite{li2022diffusion,gong2024text}. A discrete SMILES sequence is first mapped into a continuous embedding space, where diffusion is performed, and the generated representation is then decoded back into discrete SMILES tokens. Together, these components form the pretrained molecular DLM used as the generative backbone of our framework.

\subsection{SMILES Representation}
\label{sec:smiles_representation}

\noindent\textbf{Data Augmentation.}
During training, we augment each molecule by replacing its canonical SMILES string with an equivalent randomized representation generated using RDKit, following \cite{schwaller2019molecular}. This exposes the model to multiple valid representations of the same molecule and improves the diversity and robustness of molecular generation.

\noindent\textbf{Encoding and Embedding.}
Since diffusion models operate in continuous space, a SMILES sequence
\(s=(s_1,\ldots,s_n)\) is first tokenized using a Chem-Char tokenizer with vocabulary size \(v\). Each token is then projected by an embedding layer into a \(d\)-dimensional continuous representation. For a sequence of length \(n\), the resulting molecular representation is denoted as
\begin{equation}
    x_0 \in \mathbb{R}^{n \times d}.
\end{equation}
The following diffusion process is performed over this continuous molecular representation.

\subsection{Diffusion Language Model (DLM)}
\label{sec:diffusion_model}

\noindent\textbf{Forward Process.}
Given a clean molecular representation \(x_0 \sim q(x_0)\), the forward diffusion process gradually injects Gaussian noise over \(T\) steps:
\begin{equation}
\begin{array}{l}
    q(x_{1:T} \mid x_0)
    := \prod_{t=1}^{T} q(x_t \mid x_{t-1}), \\[5pt]
    q(x_t \mid x_{t-1})
    := \mathcal{N}\left(
    x_t;
    \sqrt{\alpha_t}x_{t-1},
    (1-\alpha_t)\mathbf{I}
    \right),
\end{array}
\label{eq:forward_process}
\end{equation}
where \(\alpha_t=1-\beta_t\) and \(\beta_t\) is a predefined variance schedule. As \(t\) increases, the molecular representation is progressively corrupted until
\(x_T\) approximately follows \(\mathcal{N}(0,\mathbf{I})\).

\noindent\textbf{Backward Process.}
Starting from \(x_T\sim\mathcal{N}(0,\mathbf{I})\), the reverse process progressively denoises the representation. Following the \(x_0\)-prediction formulation, a Transformer-based model predicts
\begin{equation}
    \hat{x}_0=f_\theta(x_t,t).
\end{equation}
The reverse transition is written compactly as
\begin{equation}
p_\theta(x_{t-1}\mid x_t)
=
\mathcal{N}\left(
x_{t-1};
\mu_\theta(x_t,t),
\tilde{\beta}_t\mathbf{I}
\right),
\label{eq:backward_process}
\end{equation}
where
\begin{equation}
\mu_\theta(x_t,t)
=
\frac{\sqrt{\bar{\alpha}_{t-1}}\beta_t}
{1-\bar{\alpha}_t}\hat{x}_0
+
\frac{\sqrt{\alpha_t}(1-\bar{\alpha}_{t-1})}
{1-\bar{\alpha}_t}x_t,
\end{equation}
\(
\bar{\alpha}_t=\prod_{i=1}^{t}\alpha_i
\),
and
\(
\tilde{\beta}_t=
\frac{1-\bar{\alpha}_{t-1}}
{1-\bar{\alpha}_t}\beta_t
\).
Applying this transition iteratively produces the final clean molecular representation \(x_0\).

\noindent\textbf{Rounding and Decoding.}
The generated continuous representation is finally mapped back to discrete SMILES tokens. At each token position, the \(d\)-dimensional output is mapped to its nearest token embedding using L2 distance, and the corresponding token IDs are decoded by the Chem-Char tokenizer.

\subsection{Task Formulation}
\label{sec:task_formulation}

The above encoding, diffusion, and decoding processes constitute the pretrained molecular DLM used as the generative backbone of our framework. Building on this model, our goal is to enable controllable generation under heterogeneous constraints without retraining the underlying DLM. Specifically, we consider a structural condition \(c_s\), specifying a desired molecular scaffold, and a property condition \(c_p\), specifying one or multiple desired molecular properties. Our objective is to generate
\begin{equation}
    x_0 \sim p(x_0 \mid c_s,c_p),
\end{equation}
such that the generated molecule satisfies the structural constraint while achieving the desired properties. To this end, we introduce the Structure Control Module (SCM) and Property Control Module (PCM) in the following section to provide structural and property guidance, respectively.

\section{Method}

To realize the controllable generation objective defined above, we propose CMCM-DLM, which introduces SCM and PCM on top of the pretrained molecular DLM while keeping the generative backbone frozen. As illustrated in Fig.~\ref{fig:framework}, CMCM-DLM follows a two-phase inference process: Phase I uses SCM to establish the desired structural constraint, while Phase II jointly incorporates SCM and PCM to preserve the structure and optimize the desired molecular properties.

\subsection{Phase I: Single-Modality Control}
\label{sec:scm}

Phase I aims to establish the desired molecular structure before introducing property constraints. As in the pretrained molecular DLM described in Section~\ref{sec:preliminary}, generation starts from Gaussian noise
\(x_T \sim \mathcal{N}(0,\mathbf{I})\).
Instead of unconditional denoising, Phase I incorporates structural information through the Structure Control Module (SCM). We denote the diffusion state in this phase as \(x'_t\) and the structural condition as \(c_s\).

\noindent\textbf{Structure Control Module.}
Our SCM is inspired by T2I-Adapter~\cite{mou2024t2i} and ControlNet~\cite{zhang2023control}, which introduce trainable conditioning components while keeping the pretrained diffusion backbone frozen. Following this principle, SCM incorporates structural constraints without retraining the original molecular DLM.

Given a molecular scaffold \(s\), we represent it as a SMILES sequence and encode it using a frozen pretrained scaffold encoder~\cite{Sun2024}, producing
\begin{equation}
    c_s \in \mathbb{R}^{m \times d},
\end{equation}
where \(m\) and \(d\) denote the scaffold length and embedding dimension, respectively.

As illustrated in Fig.~\ref{fig:framework}(b), SCM contains two parallel branches: a frozen pretrained backbone
\(f_{\theta_0}(x'_t,t)\), which preserves the pretrained molecular prior, and a trainable scaffold-conditioned branch
\(f_{\theta_c}(x'_t,t,c_s)\), which incorporates the scaffold condition. Their predictions are combined as
\begin{equation}
\begin{aligned}
\hat{x}_{0,s}(x'_t,t,c_s)
=\;&
w_s f_{\theta_0}(x'_t,t) \\
&+
(1-w_s)f_{\theta_c}(x'_t,t,c_s),
\end{aligned}
\label{eq:scm_equation}
\end{equation}
where \(w_s\) balances the pretrained molecular prior and structural guidance. The resulting
\(\hat{x}_{0,s}\) represents the structure-guided clean-sample prediction used during Phase I.

\noindent\textbf{Frozen Backbone.}
The frozen branch follows the Transformer architecture of the pretrained molecular DLM. It combines the noisy molecular representation \(x'_t\) with timestep and positional embeddings to construct the initial hidden representation \(z_t^{(0)}\). For the \(i\)-th layer of the \(N\)-layer Transformer,
\begin{equation}
\begin{aligned}
z_t'^{(i)}
&=
\mathrm{LN}_{\theta_0}
\left(
z_t^{(i)}
+
\mathrm{SA}_{\theta_0}(z_t^{(i)})
\right), \\
z_t^{(i+1)}
&=
\mathrm{LN}_{\theta_0}
\left(
z_t'^{(i)}
+
\mathrm{DP}_{\theta_0}
\left(
\mathrm{FFN}_{\theta_0}(z_t'^{(i)})
\right)
\right),
\end{aligned}
\label{eq:transformer_update}
\end{equation}
where \(\mathrm{SA}\), \(\mathrm{LN}\), \(\mathrm{FFN}\), and \(\mathrm{DP}\) denote self-attention, layer normalization, the feed-forward network, and dropout, respectively. After \(N\) Transformer layers, the final hidden representation \(z_t^{(N)}\) is mapped through the pretrained output head to obtain the clean-sample prediction
\(f_{\theta_0}(x'_t,t)\).

\begin{figure}[t]
    \centering
    \includegraphics[width=0.8\columnwidth]{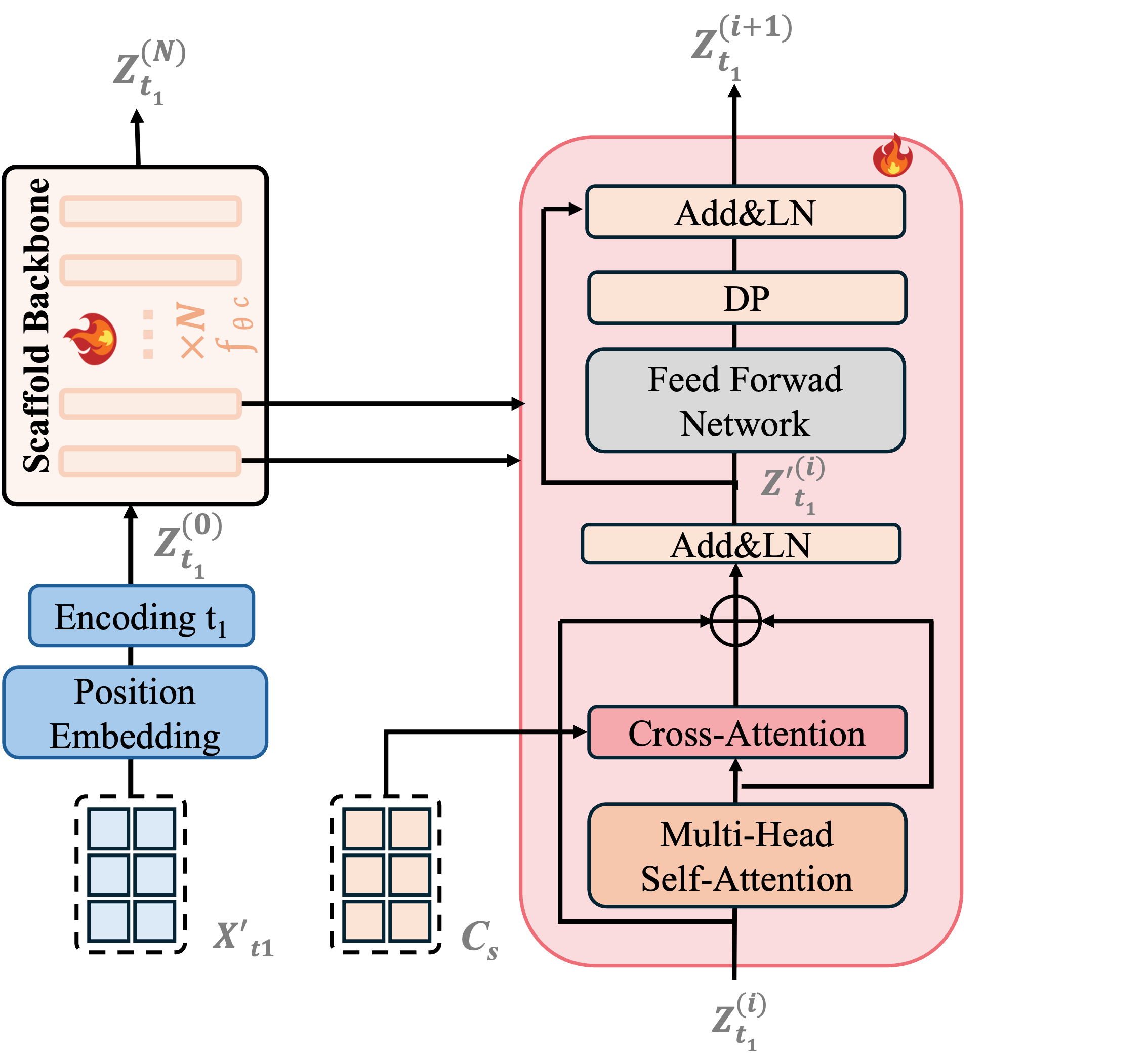}
    \vspace{-4mm}
    \caption{Architecture of the scaffold-conditioned branch
    \(f_{\theta_c}(x'_t,t,c_s)\) in SCM. Cross-attention is introduced at each Transformer layer to incorporate the scaffold representation \(c_s\).}
    \label{fig:scm_architecture}
    \vspace{-4mm}
\end{figure}

\noindent\textbf{Scaffold Backbone.}
The trainable branch follows the same Transformer architecture as the frozen backbone, while additionally incorporating the scaffold representation \(c_s\) through cross-attention at each layer. Its detailed architecture is illustrated in Fig.~\ref{fig:scm_architecture}. We denote its hidden states by \(Z_t^{(i)}\), with
\(Z_t^{(0)}=z_t^{(0)}\). The \(i\)-th layer is computed as
\begin{equation}
\begin{aligned}
Z_t'^{(i)}
&=
\mathrm{LN}_{\theta_c}
\Big(
Z_t^{(i)}
+
\mathrm{SA}_{\theta_c}(Z_t^{(i)}) \\
&+
\mathrm{CA}_{\theta_c}
\big(
\mathrm{SA}_{\theta_c}(Z_t^{(i)}),c_s
\big)
\Big), \\
Z_t^{(i+1)}
&=
\mathrm{LN}_{\theta_c}
\Big(
Z_t'^{(i)}
+
\mathrm{DP}_{\theta_c}
\big(
\mathrm{FFN}_{\theta_c}(Z_t'^{(i)})
\big)
\Big),
\end{aligned}
\label{eq:scaffold_transformer_update}
\end{equation}
where \(\mathrm{CA}\) denotes cross-attention.

After \(N\) layers, \(Z_t^{(N)}\) is mapped through the conditional output head to obtain
\(f_{\theta_c}(x'_t,t,c_s)\). Together with the frozen prediction
\(f_{\theta_0}(x'_t,t)\), \eqref{eq:scm_equation} produces the structure-guided prediction
\(\hat{x}_{0,s}(x'_t,t,c_s)\).

During Phase I, \(\hat{x}_{0,s}(x'_t,t,c_s)\) replaces the original clean-sample prediction
\(f_{\theta}(x_t,t)\) in \eqref{eq:backward_process}. The reverse diffusion process therefore progressively incorporates the desired scaffold information while retaining the molecular prior learned by the pretrained DLM. Only the scaffold-conditioned branch is optimized during SCM training, while the pretrained molecular DLM and scaffold encoder remain frozen.

\subsection{Phase II: Cross-Modality Control}
\label{sec:cross_modality}

After Phase I, the intermediate diffusion state has incorporated the desired scaffold information. Phase II further introduces property guidance while retaining the structural constraint established by SCM. We denote the diffusion state in this phase as \(x''_t\), the structural condition as \(c_s\), and the desired property conditions as
\(\mathbf{c}_p=\{c_{p,1},\ldots,c_{p,K}\}\).
As illustrated in Fig.~\ref{fig:framework}(a), SCM continues to provide structural guidance, while the Property Control Module (PCM) steers the generation toward the desired molecular properties. The two modalities are unified in the score space and jointly guide the remaining reverse diffusion steps.

\noindent\textbf{Property Control Module.}
PCM adopts classifier-based guidance~\cite{dhariwal2021diffusion} to introduce property-specific signals into the diffusion process. As shown in Fig.~\ref{fig:framework}(c), PCM contains \(K\) property predictors, where the \(k\)-th predictor models
\begin{equation}
p_{\phi_k}
\left(
c_{p,k}
\mid
x''_t,t,c_s,\mathbf{c}_{p\setminus k}
\right),
\end{equation}
where \(\phi_k\) denotes its parameters and
\(\mathbf{c}_{p\setminus k}\) contains the remaining property conditions. Conditioning on \(c_s\) and other properties allows each predictor to account for structural constraints and inter-property dependencies.

For each target property, the predictor signal is backpropagated to the current noisy molecular representation \(x''_t\), yielding
\begin{equation}
g_k
=
\nabla_{x''_t}
\log
p_{\phi_k}
\left(
c_{p,k}
\mid
x''_t,t,c_s,\mathbf{c}_{p\setminus k}
\right).
\label{eq:property_gradient_single}
\end{equation}
For multiple properties, these gradients are aggregated as
\begin{equation}
g_p(x''_t,t,c_s,\mathbf{c}_p)
=
\sum_{k=1}^{K}\lambda_k g_k,
\label{eq:property_gradient}
\end{equation}
where \(\lambda_k\) controls the contribution of each property. Thus, \(g_p\) provides a property-specific direction in the noisy molecular space toward the desired property profile.

\noindent\textbf{Score Fusion for Cross-Modality Control.}
To combine structural and property guidance, we express both modalities in a common score space following score-based diffusion modeling~\cite{song2021score}. For an \(x_0\)-prediction model \(f_\theta\), the corresponding score at diffusion state \(x''_t\) is
\begin{equation}
s_\theta(x''_t,t)
=
\frac{
\sqrt{\bar{\alpha}_t}f_\theta(x''_t,t)-x''_t
}{
1-\bar{\alpha}_t
}.
\label{eq:score_conversion}
\end{equation}
Accordingly, the frozen backbone \(f_{\theta_0}(x''_t,t)\) yields the molecular prior score \(s_0(x''_t,t)\). Similarly, applying the same conversion to the structure-guided prediction \(\hat{x}_{0,s}(x''_t,t,c_s)\) produced by SCM yields the structure-guided score \(s_s(x''_t,t,c_s)\).

PCM combines the molecular prior with the aggregated property gradient \(g_p\) to obtain the property-guided score:
\begin{equation}
\tilde{s}_p
=
w_p\,s_0(x''_t,t)
+
(1-w_p)\,
g_p(x''_t,t,c_s,\mathbf{c}_p),
\label{eq:pcm_equation}
\end{equation}
where \(w_p\) balances the pretrained molecular prior and property guidance.

The structure- and property-guided scores are then additively fused:
\begin{equation}
\tilde{s}
(x''_t,t,c_s,\mathbf{c}_p)
=
s_s(x''_t,t,c_s)
+
\tilde{s}_p(x''_t,t,c_s,\mathbf{c}_p).
\label{eq:over_equation}
\end{equation}
The fused score is converted back to the corresponding \(x_0\)-prediction through the inverse of \eqref{eq:score_conversion} and used in the remaining reverse diffusion steps. In this way, Phase II jointly incorporates structural and property constraints to generate
\(
x_0 \sim p(x_0\mid c_s,\mathbf{c}_p)
\),
realizing the cross-modality controllable generation objective defined in Section~\ref{sec:task_formulation}. A theoretical analysis of \eqref{eq:over_equation} is provided in Section~\ref{sec:theory}.

\subsection{Training and Inference}
\label{sec:training_inference}

We train SCM and PCM separately on top of the frozen pretrained molecular DLM, and combine them only during the two-phase inference process described above.

\noindent\textbf{SCM Training.}
SCM is trained on molecular representations paired with their scaffold conditions. Given a clean representation \(x_0\), we sample \(x_t\) using the forward diffusion process and optimize the scaffold-conditioned branch to reconstruct \(x_0\):
\begin{equation}
\mathcal{L}_{\mathrm{SCM}}
=
\mathbb{E}
\left[
\left\|
f_{\theta_c}(x_t,t,c_s)-x_0
\right\|_2^2
-\log r_\theta(x\mid x_0)
\right],
\label{eq:scm_loss}
\end{equation}
where \(r_\theta(x\mid x_0)\) denotes the rounding process. The pretrained molecular DLM and scaffold encoder remain frozen during training.

\noindent\textbf{PCM Training.}
For PCM, we train Transformer-based property predictors with 12 layers and 12 attention heads. Each predictor receives a noisy molecular representation \(x_t\), the timestep \(t\), and, when available, the scaffold and other property conditions. The \(K\) predictors are optimized using
\begin{equation}
\mathcal{L}_{\mathrm{PCM}}
=
\mathbb{E}
\left[
\frac{1}{K}
\sum_{k=1}^{K}
\left\|
\hat{c}_{p,k}-c_{p,k}
\right\|_2^2
\right],
\label{eq:pcm_loss}
\end{equation}
where
\(
\hat{c}_{p,k}
=
p_{\phi_k}(x_t,t,c_s,\mathbf{c}_{p\setminus k})
\).
During inference, gradients of these predictors with respect to \(x_t\) provide the property guidance described in \eqref{eq:property_gradient}.

\noindent\textbf{Inference.}
CMCM-DLM performs reverse diffusion in two phases separated by a transition timestep \(\tau\). Phase I applies SCM from \(T\) to \(\tau\) to establish the desired scaffold, while Phase II jointly applies SCM and PCM from \(\tau\) to \(0\) for cross-modality control. The complete sampling procedure is summarized in Algorithm~\ref{alg:conditional_sampling}.

\begin{algorithm}[t]
\caption{Inference of CMCM-DLM}
\label{alg:conditional_sampling}
\small
\begin{algorithmic}[1]
\Require Structural condition \(c_s\), property conditions \(\mathbf{c}_p\), transition timestep \(\tau\)
\State Sample \(x'_T \sim \mathcal{N}(0,\mathbf{I})\)

\For{\(t=T,\ldots,\tau+1\)} \Comment{Phase I}
    \State Compute \(\hat{x}_{0,s}(x'_t,t,c_s)\) using \eqref{eq:scm_equation}
    \State Sample \(x'_{t-1}\) using \eqref{eq:backward_process} 
\EndFor

\State \(x''_{\tau} \gets x'_{\tau}\)

\For{\(t=\tau,\ldots,1\)} \Comment{Phase II}
    \State Compute property gradients \(g_p\) using \eqref{eq:property_gradient}
    \State Compute \(\hat{x}_{0,s}(x''_t,t,c_s)\) using \eqref{eq:scm_equation}
    \State Convert \(f_{\theta_0}(x''_t,t)\) and \(\hat{x}_{0,s}\) to \(s_0\) and \(s_s\) using \eqref{eq:score_conversion}
    \State Compute \(\tilde{s}_p\) using \eqref{eq:pcm_equation}
    \State Fuse \(\tilde{s}=s_s+\tilde{s}_p\) using \eqref{eq:over_equation}
    \State Convert \(\tilde{s}\) to its corresponding clean-sample prediction \(\hat{x}_{0,\mathrm{CM}}\)
    \State Sample \(x''_{t-1}\) using \eqref{eq:backward_process} 
\EndFor

\Ensure Decode \(x_0\) into the final SMILES sequence
\end{algorithmic}
\end{algorithm}

\begin{table*}[t]
\centering
\caption{Evaluation of controllable molecular generation methods on the GuacaMol, ZINC, and QM9 datasets. For CMCM-DLM, QED, SAS, and PLogP are jointly optimized under the scaffold constraint. ``Improvement (\%)'' denotes the relative property improvement over the mean property values of training molecules containing the corresponding scaffold, using the absolute baseline value in the denominator. Higher QED and PLogP and lower SAS indicate better performance. The reported improvements correspond to QED, SAS, and PLogP, respectively. Scaffold membership is determined using RDKit's ScaffoldExistence function.}
\label{tab:over_metrics}

\resizebox{\textwidth}{!}{%
\begin{tabular}{@{}llcccccccc@{}}
\toprule

\multicolumn{2}{c}{} &
\multicolumn{2}{c}{\textbf{Basic}} &
\multicolumn{4}{c}{\textbf{Property}} &
\multicolumn{2}{c}{\textbf{Structure}} \\
\cmidrule(lr){3-4}
\cmidrule(lr){5-8}
\cmidrule(lr){9-10}

\textbf{Dataset} &
\textbf{Method} &
\textbf{Validity} $\uparrow$ &
\textbf{Novelty} $\uparrow$ &
\textbf{QED} $\uparrow$ &
\textbf{SAS} $\downarrow$ &
\textbf{PLogP} $\uparrow$ &
\textbf{Improvement (\%)} $\uparrow$ &
\textbf{Scaffold Similarity (\%)} $\uparrow$ &
\textbf{Scaffold Existence (\%)} $\uparrow$ \\

\midrule

GuacaMol &
TGM-DLM &
97\% & 99\% & 0.573 & 2.622 & -7.013 &
[8\%, 5\%, -1\%] & 35\% & 54\% \\

&
MolGPT &
89\% & 94\% & 0.642 & 2.343 & -6.872 &
[21\%, 15\%, 1\%] & 18\% & 42\% \\

&
ChemFM &
84\% & 96\% & \textbf{0.701} & 2.276 & -7.366 &
[\textbf{32\%}, 18\%, -6\%] & 33\% & 39\% \\

&
\textbf{CMCM-DLM (Ours)} &
85\% & 97\% & 0.697 & \textbf{2.124} & \textbf{-6.742} &
[32\%, \textbf{23\%}, \textbf{3\%}] &
\textbf{55\%} & \textbf{88\%} \\

\midrule

ZINC &
TGM-DLM &
96\% & 98\% & 0.668 & 2.450 & -4.413 &
[-3\%, 14\%, 5\%] & 34\% & 52\% \\

&
MolGPT &
90\% & 95\% & 0.707 & 2.615 & -4.872 &
[2\%, 9\%, -5\%] & 46\% & 51\% \\

&
ChemFM &
93\% & 97\% & 0.729 & 2.492 & -4.287 &
[6\%, 13\%, 7\%] & 49\% & 62\% \\

&
\textbf{CMCM-DLM (Ours)} &
85\% & 94\% & \textbf{0.742} & \textbf{2.433} & \textbf{-4.202} &
[\textbf{7\%}, \textbf{15\%}, \textbf{9\%}] &
\textbf{52\%} & \textbf{83\%} \\

\midrule

QM9 &
TGM-DLM &
94\% & 96\% & 0.558 & 2.962 & -1.131 &
[12\%, 1\%, -28\%] & 32\% & 48\% \\

&
MolGPT &
86\% & 91\% & 0.548 & 2.856 & -0.966 &
[10\%, 5\%, -9\%] & 43\% & 65\% \\

&
ChemFM &
89\% & 93\% & 0.554 & 2.579 & \textbf{-0.802} &
[11\%, 14\%, \textbf{9\%}] & 49\% & 71\% \\

&
\textbf{CMCM-DLM (Ours)} &
86\% & 93\% & \textbf{0.571} & \textbf{2.546} & -0.850 &
[\textbf{14\%}, \textbf{15\%}, 4\%] &
\textbf{63\%} & \textbf{78\%} \\

\bottomrule
\end{tabular}%
}

\end{table*}

\section{Theoretical Analysis}
\label{sec:theory}

We provide a theoretical interpretation of the cross-modality score 
fusion in \eqref{eq:over_equation}. By expressing structural and 
property guidance in a common score space, CMCM-DLM combines 
heterogeneous constraints while preserving the molecular prior of the 
pretrained DLM.

\subsection{Cross-Modality Score Fusion}
\label{sec:proof}

We begin with the conditional distribution
\begin{equation}
p(x''_t \mid t,c_s,\mathbf{c}_p)
\propto
p(x''_t\mid t)\,
p(c_s,\mathbf{c}_p\mid x''_t,t).
\end{equation}
Using the chain rule,
\begin{equation}
p(c_s,\mathbf{c}_p\mid x''_t,t)
=
p(c_s\mid x''_t,t)\,
p(\mathbf{c}_p\mid x''_t,t,c_s).
\end{equation}

To balance structural and property guidance, we introduce a weight
\(w\) and approximate the conditional distribution as a weighted
composition of the two modality-specific constraints:
\begin{equation}
\begin{aligned}
\log p(x''_t\mid t,c_s,\mathbf{c}_p)
\approx\;&
\log p(x''_t\mid t) \\
&+
w\log p(c_s\mid x''_t,t) \\
&+
(1-w)\log p(\mathbf{c}_p\mid x''_t,t,c_s).
\end{aligned}
\label{eq:proof_log}
\end{equation}

Taking the gradient with respect to \(x''_t\), we obtain
\begin{equation}
\begin{aligned}
\tilde{s}
\approx\;&
s_0(x''_t,t) \\
&+
w\nabla_{x''_t}\log p(c_s\mid x''_t,t) \\
&+
(1-w)\nabla_{x''_t}
\log p(\mathbf{c}_p\mid x''_t,t,c_s),
\end{aligned}
\label{eq:proof_gradient}
\end{equation}
where
\(
s_0(x''_t,t)=\nabla_{x''_t}\log p(x''_t\mid t)
\)
denotes the pretrained molecular prior score.

In CMCM-DLM, the two guidance terms are instantiated by SCM and PCM,
respectively. The property-dependent term is realized by PCM through
the aggregated property gradient
\(g_p(x''_t,t,c_s,\mathbf{c}_p)\) defined in
\eqref{eq:property_gradient}. Accordingly, the property-guided score is
written as
\begin{equation}
\tilde{s}_p
=
w_p s_0(x''_t,t)
+
(1-w_p)
g_p(x''_t,t,c_s,\mathbf{c}_p),
\label{eq:theory_pcm}
\end{equation}
where \(w_p\) controls the balance between the molecular prior and
property guidance.

For structural guidance, SCM learns a scaffold-conditioned diffusion
branch. Following \eqref{eq:score_conversion}, its corresponding score is
\begin{equation}
s_c(x''_t,t,c_s)
=
\frac{
\sqrt{\bar{\alpha}_t}
f_{\theta_c}(x''_t,t,c_s)-x''_t
}{
1-\bar{\alpha}_t
}.
\end{equation}

Since \eqref{eq:scm_equation} linearly combines the frozen and
scaffold-conditioned \(x_0\)-predictions, the corresponding
structure-guided score is
\begin{equation}
s_s(x''_t,t,c_s)
=
w_s s_0(x''_t,t)
+
(1-w_s)s_c(x''_t,t,c_s).
\label{eq:theory_scm}
\end{equation}

To obtain a unified decomposition of the prior score, we assign
complementary weights to the two modalities:
\(
w_s=1-w_p
\).
The structural and property guidance can therefore be jointly composed
as
\begin{equation}
\boxed{
\tilde{s}(x''_t,t,c_s,\mathbf{c}_p)
\approx
s_s(x''_t,t,c_s)
+
\tilde{s}_p(x''_t,t,c_s,\mathbf{c}_p)
}
\label{eq:theory_fusion}
\end{equation}
which corresponds to \eqref{eq:over_equation}. This formulation
retains the molecular prior learned by the pretrained DLM while
allowing structural and property constraints to jointly guide the
reverse diffusion process in a unified score space.

\section{Experiments}
We evaluate CMCM-DLM from three perspectives: (1) its ability to jointly handle heterogeneous structural and property constraints, (2) the effectiveness of SCM and PCM for their respective control modalities, and (3) the flexibility and efficiency of adapting a pretrained molecular DLM to different control requirements. Experiments are conducted across molecular datasets with diverse scales, molecular distributions, and property spaces.
\subsection{Experimental setup}

\noindent\textbf{Datasets.}
To evaluate CMCM-DLM under heterogeneous molecular data and control requirements, we conduct experiments on three widely used datasets with different scales, molecular distributions, and property spaces: GuacaMol, ZINC250K, and QM9. GuacaMol~\cite{brown2019guacamol}, derived from ChEMBL 24~\cite{mendez2019chembl}, contains approximately 1.6 million bioactive molecules with diverse chemical structures and properties, providing a large-scale setting for controlled molecular generation. ZINC250K~\cite{irwin2012zinc} contains 249,455 drug-like molecules and covers a broad range of molecular structures, making it suitable for evaluating both structural and property control. QM9~\cite{ramakrishnan2014quantum} consists of 133,885 small organic molecules with up to nine heavy atoms and provides a diverse set of quantum-chemical properties computed using density functional theory (DFT). Together, these datasets span different data scales, molecular domains, and property distributions, enabling us to assess the robustness and adaptability of CMCM-DLM across heterogeneous molecular generation scenarios.

\begin{table*}[!t]
\centering
\caption{Additional cross-modality control results of CMCM-DLM on
GuacaMol, ZINC, and QM9 under single- and pairwise-property
constraints. ``Improvement (\%)'' reports the relative improvement of
the optimized properties in the order listed in each setting. All
results are evaluated on 10,000 generated molecules.}
\label{tab:additional_cross_modality}

\resizebox{\textwidth}{!}{%
\begin{tabular}{@{}llcccccccc@{}}
\toprule

\multicolumn{2}{c}{} &
\multicolumn{2}{c}{\textbf{Basic}} &
\multicolumn{4}{c}{\textbf{Property}} &
\multicolumn{2}{c}{\textbf{Structure}} \\
\cmidrule(lr){3-4}
\cmidrule(lr){5-8}
\cmidrule(lr){9-10}

\textbf{Dataset} &
\textbf{Method} &
\textbf{Validity} $\uparrow$ &
\textbf{Novelty} $\uparrow$ &
\textbf{QED} $\uparrow$ &
\textbf{SAS} $\downarrow$ &
\textbf{PLogP} $\uparrow$ &
\textbf{Improvement (\%)} &
\textbf{Scaffold Similarity (\%)} $\uparrow$ &
\textbf{Scaffold Existence (\%)} $\uparrow$ \\

\midrule

GuacaMol &
CMCM-DLM$_{\text{QED}}$ &
80\% & 100\% & 0.710 & 2.610 & -7.890 &
[34\%] & 51\% & 82\% \\

&
CMCM-DLM$_{\text{SAS}}$ &
86\% & 97\% & 0.670 & 2.020 & -6.690 &
[27\%] & 55\% & 85\% \\

&
CMCM-DLM$_{\text{PLogP}}$ &
90\% & 99\% & 0.530 & 2.100 & -5.420 &
[22\%] & 60\% & 87\% \\

&
CMCM-DLM$_{\text{QED+SAS}}$ &
83\% & 98\% & 0.700 & 2.220 & -7.240 &
[32\%, 20\%] & 52\% & 84\% \\

&
CMCM-DLM$_{\text{QED+PLogP}}$ &
82\% & 100\% & 0.690 & 2.300 & -6.800 &
[30\%, 2\%] & 52\% & 84\% \\

&
CMCM-DLM$_{\text{SAS+PLogP}}$ &
82\% & 98\% & 0.670 & 2.130 & -6.690 &
[23\%, 4\%] & 53\% & 82\% \\

\midrule

ZINC &
CMCM-DLM$_{\text{QED}}$ &
92\% & 98\% & 0.780 & 2.830 & -6.320 &
[13\%] & 48\% & 76\% \\

&
CMCM-DLM$_{\text{SAS}}$ &
93\% & 93\% & 0.620 & 2.320 & -4.430 &
[19\%] & 53\% & 82\% \\

&
CMCM-DLM$_{\text{PLogP}}$ &
84\% & 99\% & 0.560 & 2.410 & -4.120 &
[11\%] & 61\% & 76\% \\

&
CMCM-DLM$_{\text{QED+SAS}}$ &
88\% & 97\% & 0.750 & 2.450 & -5.310 &
[9\%, 14\%] & 57\% & 65\% \\

&
CMCM-DLM$_{\text{QED+PLogP}}$ &
84\% & 100\% & 0.730 & 2.520 & -4.280 &
[6\%, 8\%] & 44\% & 72\% \\

&
CMCM-DLM$_{\text{SAS+PLogP}}$ &
91\% & 98\% & 0.560 & 2.390 & -4.140 &
[17\%, 11\%] & 56\% & 84\% \\

\midrule

QM9 &
CMCM-DLM$_{\text{QED}}$ &
82\% & 99\% & 0.620 & 3.130 & -3.210 &
[24\%] & 61\% & 83\% \\

&
CMCM-DLM$_{\text{SAS}}$ &
81\% & 98\% & 0.510 & 2.460 & -2.390 &
[18\%] & 54\% & 78\% \\

&
CMCM-DLM$_{\text{PLogP}}$ &
78\% & 98\% & 0.460 & 2.720 & -0.750 &
[15\%] & 60\% & 72\% \\

&
CMCM-DLM$_{\text{QED+SAS}}$ &
85\% & 99\% & 0.610 & 2.520 & -2.890 &
[22\%, 16\%] & 63\% & 76\% \\

&
CMCM-DLM$_{\text{QED+PLogP}}$ &
74\% & 100\% & 0.570 & 2.930 & -0.820 &
[14\%, 7\%] & 56\% & 68\% \\

&
CMCM-DLM$_{\text{SAS+PLogP}}$ &
83\% & 99\% & 0.490 & 2.510 & -0.760 &
[16\%, 14\%] & 64\% & 75\% \\

\bottomrule
\end{tabular}%
}

\end{table*}

\noindent\textbf{Scaffold Evaluation Set.}
\label{sec:vs_set}
we construct a Scaffold Evaluation Set (VS Set) containing 10 representative molecular scaffolds. The selected scaffolds cover diverse chemical structures and appear frequently across the molecular datasets used in our experiments. Specifically, the VS Set includes Piperazine (\texttt{N1CCNCC1}),
Pyrimidine (\texttt{c1cncnc1}), Pyrazine (\texttt{c1cnccn1}),
Pyridine (\texttt{c1ccncc1}), Benzene (\texttt{c1ccccc1}), Furan
(\texttt{c1ccoc1}), Phenol (\texttt{c1ccccc1O}), Benzothiazole
(\texttt{c1ccc2ncsc2c1}), Thiazole (\texttt{c1cscn1}), and Naphthalene
(\texttt{c1ccc2ccccc2c1}). These scaffolds occur frequently across the
molecular datasets used in our experiments, with an average occurrence
rate of 17.6\%, providing a representative evaluation set for
assessing scaffold preservation and structural control.

\noindent\textbf{Evaluation:}
The generated molecules were evaluated using three core metrics:
validity, novelty, and diversity. Validity is the percentage of
chemically valid molecules, assessed with RDKit~\cite{rdkit}.
Novelty measures the fraction of valid molecules not appearing in the
training set, reflecting generative originality. Diversity assesses
the uniqueness among valid molecules, indicating sample variability.

Secondly, we evaluate molecular properties using QED, SAS, PLogP,
HBA, and HBD. QED~\cite{bickerton2012quantifying} measures
drug-likeness on a scale from 0 to 1, while SAS~\cite{ertl2009sas}
evaluates synthetic accessibility from 1 (easy) to 10 (difficult).
PLogP is the penalized logP score, which combines lipophilicity with
synthetic accessibility and ring penalties:
\begin{equation}
\begin{aligned}
\mathrm{PLogP}
=\;&
\frac{\mathrm{LogP}-\mu_{\mathrm{LogP}}}
{\sigma_{\mathrm{LogP}}}
+
\frac{\mathrm{SAS}-\mu_{\mathrm{SAS}}}
{\sigma_{\mathrm{SAS}}}
\\
&+
\frac{\mathrm{CyclePenalty}-\mu_{\mathrm{Cycle}}}
{\sigma_{\mathrm{Cycle}}}.
\end{aligned}
\end{equation}
Higher QED and PLogP values indicate better molecular properties,
whereas lower SAS values indicate improved synthetic accessibility.
Finally, HBA and HBD denote the numbers of hydrogen-bond acceptors
and donors, respectively, which are commonly used in drug-likeness
assessment and are closely associated with molecular absorption and
permeability~\cite{lipinski1997experimental}.

Lastly, we assessed scaffold fidelity using two metrics: scaffold
existence and scaffold similarity. Scaffold existence is determined
via RDKit’s HasSubstructMatch, which checks whether a generated
molecule contains the target scaffold. Scaffold similarity quantifies
structural resemblance by computing the cosine similarity (0--1)
between MACCS key fingerprints of the molecule and the scaffold
~\cite{durant2002reoptimization}.

\noindent\textbf{Implementation Details.}
All experiments are conducted on NVIDIA V100 32GB GPUs. The pretrained molecular DLM consists of 12 Transformer layers with 12 attention heads and a hidden dimension of 768, using a maximum SMILES length of 100, a vocabulary size of 590, and \(T=2000\) diffusion steps. The backbone is trained with Adam at a learning rate of \(1\times10^{-4}\). SCM follows the backbone architecture with cross-attention added at each layer, while PCM uses a 12-layer, 12-head Transformer property predictor trained on diffusion steps below 1500 with learning rates from \(1\times10^{-5}\) to \(1\times10^{-4}\). We set \(t=1500\) as the transition point between Phase I and Phase II.

\begin{table*}[t]
\centering
\caption{An optimization example from the GuacaMol dataset. The first column shows the scaffold, followed by the optimized molecules generated by CMCM-DLM under different property control combinations: QED+SAS+PLogP, QED+SAS, QED+PLogP, and SAS+PLogP. A check mark ($\checkmark$) indicates improvement over the scaffold baseline, which is calculated as the mean property scores of all GuacaMol molecules containing that scaffold. The property scores (QED~($\uparrow$), SAS~($\downarrow$), PLogP~($\uparrow$)) and scaffold similarity are used to evaluate constraint satisfaction and optimization quality. The gray dashed rectangle highlights the location of the retained scaffold.}
\label{tab:optimization_example}

\footnotesize
\setlength{\tabcolsep}{2.5pt}

\resizebox{\textwidth}{!}{%
\begin{tabular}{@{\extracolsep{\fill}}lccccc@{}}
\toprule

& \textbf{Scaffold}
& \textbf{CMCM-DLM\textsubscript{QED+SAS+PLOGP}}
& \textbf{CMCM-DLM\textsubscript{QED+SAS}}
& \textbf{CMCM-DLM\textsubscript{QED+PLOGP}}
& \textbf{CMCM-DLM\textsubscript{SAS+PLOGP}}
\\

\textbf{Molecule}
&
\raisebox{-.5\height}{
    \includegraphics[height=1.2cm]{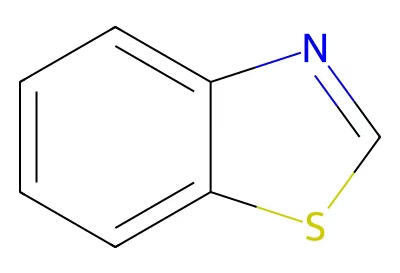}
}
&
\raisebox{-.5\height}{
    \includegraphics[height=1.1cm]{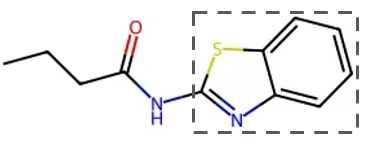}
}
&
\raisebox{-.5\height}{
    \includegraphics[height=1.1cm]{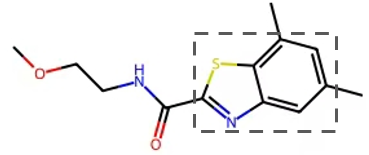}
}
&
\raisebox{-.5\height}{
    \includegraphics[height=1.1cm]{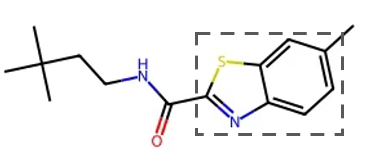}
}
&
\raisebox{-.5\height}{
    \includegraphics[height=1.1cm]{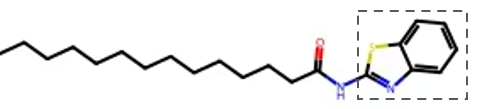}
}
\\

\midrule

\textbf{SMILES String}
&
\raisebox{-.5\height}{
    \shortstack[l]{
        \texttt{c1ccc2ncsc2c1}
    }
}
&
\raisebox{-.5\height}{
    \shortstack[l]{
        \texttt{CCCC(=O)Nc1}\\
        \texttt{nc2ccccc2s1}
    }
}
&
\raisebox{-.5\height}{
    \shortstack[l]{
        \texttt{COCCNC(=O)c1n}\\
        \texttt{c2cc(C)cc(C)c2s1}
    }
}
&
\raisebox{-.5\height}{
    \shortstack[l]{
        \texttt{CC(C)(C)CCNC(=O)c}\\
        \texttt{1nc2ccc(C)cc2s1}
    }
}
&
\raisebox{-.5\height}{
    \shortstack[l]{
        \texttt{CCCCCCCCCCCCC}\\
        \texttt{C()=ONc1nc2ccccc2s1}
    }
}
\\

\midrule

\textbf{QED ($\uparrow$)}
& 0.500
& 0.863 $\checkmark$
& 0.862 $\checkmark$
& 0.927 $\checkmark$
& 0.393
\\

\textbf{SAS ($\downarrow$)}
& 2.660
& 1.716 $\checkmark$
& 2.310 $\checkmark$
& 2.228 $\checkmark$
& 1.918 $\checkmark$
\\

\textbf{PLogP ($\uparrow$)}
& -6.360
& -5.783 $\checkmark$
& -6.928
& -6.012 $\checkmark$
& -3.888 $\checkmark$
\\

\textbf{Scaffold Similarity ($\uparrow$)}
& --
& 0.663
& 0.606
& 0.640
& 0.626
\\

\bottomrule
\end{tabular}%
}

\end{table*}

\begin{table}[t]
\centering
\caption{PCM performance on single- and multi-property optimization using GuacaMol. The symbols \(+\) and \(-\) denote increasing and decreasing HBA/HBD, respectively. Improvement (\%) measures the relative change compared with the dataset mean.}
\label{tab:guacamol_metrics}

\resizebox{\columnwidth}{!}{%
\begin{tabular}{@{}p{2.5cm}cccc@{}}
\toprule
\multicolumn{1}{c}{} & \multicolumn{4}{c}{GuacaMol} \\
\cmidrule(lr){2-5}

Method & Validity $\uparrow$ & Novelty $\uparrow$ & Diversity $\uparrow$ & Improvement $\uparrow$ \\
\midrule

PCM\textsubscript{QED} & 86\% & 99\% & 100\% & [58\%] \\
PCM\textsubscript{SAS} & 88\% & 96\% & 99\% & [26\%] \\
PCM\textsubscript{PLogP} & 82\% & 99\% & 90\% & [47\%] \\

\midrule

PCM\textsubscript{HBA$+$} & 66\% & 100\% & 100\% & [67\%] \\
PCM\textsubscript{HBA$-$} & 87\% & 97\% & 95\% & [74\%] \\
PCM\textsubscript{HBD$+$} & 93\% & 98\% & 97\% & [142\%] \\
PCM\textsubscript{HBD$-$} & 91\% & 98\% & 99\% & [91\%] \\

\midrule

PCM\textsubscript{QED+SAS} & 91\% & 98\% & 98\% & [56\%, 25\%] \\
PCM\textsubscript{QED+PLogP} & 93\% & 98\% & 97\% & [44\%, 19\%] \\
PCM\textsubscript{SAS+PLogP} & 83\% & 99\% & 91\% & [35\%, 49\%] \\

\midrule

PCM\textsubscript{QED+SAS+PLogP} & 93\% & 98\% & 97\% & [44\%, 31\%, 18\%] \\

\bottomrule
\end{tabular}%
}
\end{table}

\subsection{Performance of CMCM-DLM}
\label{sec:over_i}

We first evaluate CMCM-DLM under the most challenging cross-modality setting, where a scaffold constraint and three molecular properties (QED, SAS, and PLogP) are jointly controlled. We compare against TGM-DLM~\cite{gong2024text}, MolGPT~\cite{bagal2022molgpt}, and ChemFM~\cite{cai2026chemfm}. TGM-DLM is a text-guided diffusion language model, for which we convert the scaffold and property requirements into textual conditions for generation. MolGPT is an autoregressive Transformer-based molecular generator supporting conditional generation, while ChemFM is a large-scale chemical foundation model that can be adapted to property-conditioned molecular design. For CMCM-DLM, each SCM is trained on the same dataset as its corresponding pretrained DLM, whereas the PCM is trained only on GuacaMol and reused across datasets.

For each setting, we generate 20,000 molecules and evaluate validity, novelty, QED, SAS, PLogP, scaffold similarity, and scaffold existence. The complete scaffold list is provided in Section~\ref{sec:vs_set}. Table~\ref{tab:over_metrics} reports the joint QED+SAS+PLogP results under scaffold control.

As shown in Table~\ref{tab:over_metrics}, CMCM-DLM achieves the strongest structural control across all three datasets. Scaffold similarity reaches 55\%, 52\%, and 63\% on GuacaMol, ZINC, and QM9, respectively, while scaffold existence reaches 88\%, 83\%, and 78\%, consistently outperforming the baselines. CMCM-DLM also provides strong joint property control, achieving the best or competitive QED, SAS, and PLogP results while maintaining high novelty of 93--97\%. The relative improvements reach [32\%, 23\%, 3\%], [7\%, 15\%, 9\%], and [14\%, 15\%, 4\%] on GuacaMol, ZINC, and QM9 for QED, SAS, and PLogP, respectively.

The relatively smaller PLogP improvement under joint control reflects the competition among multiple property objectives and the scaffold constraint. For example, on GuacaMol, PLogP improves by 22\% when optimized individually under scaffold control, compared with 3\% when QED, SAS, and PLogP are optimized jointly. This indicates that CMCM-DLM can provide stronger optimization for individual objectives, while joint control requires balancing potentially competing constraints.

Additional single- and pairwise-property control results are
summarized in Table~\ref{tab:additional_cross_modality}. These results further indicate that the smaller PLogP improvement under triple-property optimization is caused by the competition among heterogeneous objectives rather than the inability of CMCM-DLM to optimize molecular properties. Under single-property control, CMCM-DLM achieves effective optimization for different objectives, including PLogP improvements of 22\%, 11\%, and 15\% on GuacaMol, ZINC, and QM9, respectively. When multiple properties are jointly optimized, the improvements are redistributed across objectives due to different optimization requirements, while other properties such as QED and SAS can still maintain substantial gains. Meanwhile, scaffold fidelity remains stable across different constraint combinations. These results demonstrate that CMCM-DLM can flexibly integrate heterogeneous molecular constraints and adapt its control capability to diverse optimization scenarios.

\begin{table}[t]
\centering
\caption{Comparison of QED scores for the top-$k$ molecules out of 10,000 generated candidates. Results of GraphNVP, GraphAF, GDSS, and MoFlow are adapted from \cite{guo2024diffusing}.}
\label{tab:qed_metrics}

\resizebox{\columnwidth}{!}{%
\begin{tabular}{@{}lccccc@{}}
\toprule
\multicolumn{1}{c}{} &
\multicolumn{5}{c}{\textbf{QED} $\uparrow$} \\
\cmidrule(lr){2-6}

\textbf{Method} &
\textbf{Top-1} &
\textbf{Top-5} &
\textbf{Top-10} &
\textbf{Top-100} &
\textbf{Top-1000} \\
\midrule

GraphNVP &
0.8512 &
0.8391 &
0.8260 &
0.7423 &
0.5046 \\

GraphAF &
0.9437 &
0.9261 &
0.9176 &
0.8593 &
0.5749 \\

GDSS &
0.9449 &
0.9369 &
0.9337 &
0.9126 &
0.8512 \\

MoFlow &
0.9261 &
0.9233 &
0.9150 &
0.8664 &
0.7839 \\

D2L-OMP &
0.9476 &
0.9417 &
0.9372 &
0.9144 &
0.8529 \\

\midrule

\textbf{CMCM-DLM (ours)} &
\textbf{0.9481} &
\textbf{0.9467} &
\textbf{0.9458} &
\textbf{0.9413} &
\textbf{0.9189} \\

\bottomrule
\end{tabular}%
}

\end{table}

\subsection{Influence of PCM} \label{sec:pcm_i}

We evaluated PCM without SCM on single- and multi-property optimization tasks using the GuacaMol dataset. As shown in Table~\ref{tab:guacamol_metrics}, PCM achieves strong property controllability while maintaining high molecular quality. For QED, SAS, and PLogP, the model achieves substantial improvements under both single- and multi-property settings. We further evaluate HBA and HBD by controlling them in both increasing and decreasing directions. PCM achieves an average improvement of \(96\%\) when increasing HBA and HBD and \(83\%\) when decreasing them, demonstrating effective control over properties without a predefined optimization direction.

Since HBA and HBD do not have an inherent optimization direction, we exclude them from the cross-modality experiments and focus on QED, SAS, and PLogP for evaluating CMCM-DLM.

To benchmark our framework against state-of-the-art methods, we conducted a comparative study on QED optimization using the ZINC dataset (Table~\ref{tab:qed_metrics}). Since the competing methods did not specify how they defined PLogP, we focused on QED optimization, where our framework consistently outperformed all competitors across various ranking thresholds. It achieved top scores of 0.9481 (Top-1), 0.9467 (Top-5), 0.9458 (Top-10), 0.9413 (Top-100), and 0.9189 (Top-1000).

\subsection{Effectiveness of SCM} \label{sec:scm_i}

We also evaluated the performance of SCM without PCM on the VS set and further tested its generalization using five additional scaffolds from SAMOA~\cite{langevin2020scaffold} as zero-shot samples, none of which appeared in the training data. On the VS set, scaffold similarity remains stable at \(60\%\), and scaffold existence reaches \(80\%-90\%\) as training progresses. For the zero-shot scaffolds, similarity improves to \(80\%-86\%\), with scaffold existence around \(59\%\), demonstrating strong generalization capability. SCM also maintains high molecule quality, achieving an average validity of \(86\%\), nearly \(100\%\) novelty, and around \(80\%\) diversity across both the VS set and zero-shot samples. Moreover, it reaches strong performance in just five epochs, requiring less than \(20\%\) of the training time of a full diffusion model.

\begin{figure}[t]
    \centering
    \includegraphics[width=\columnwidth]{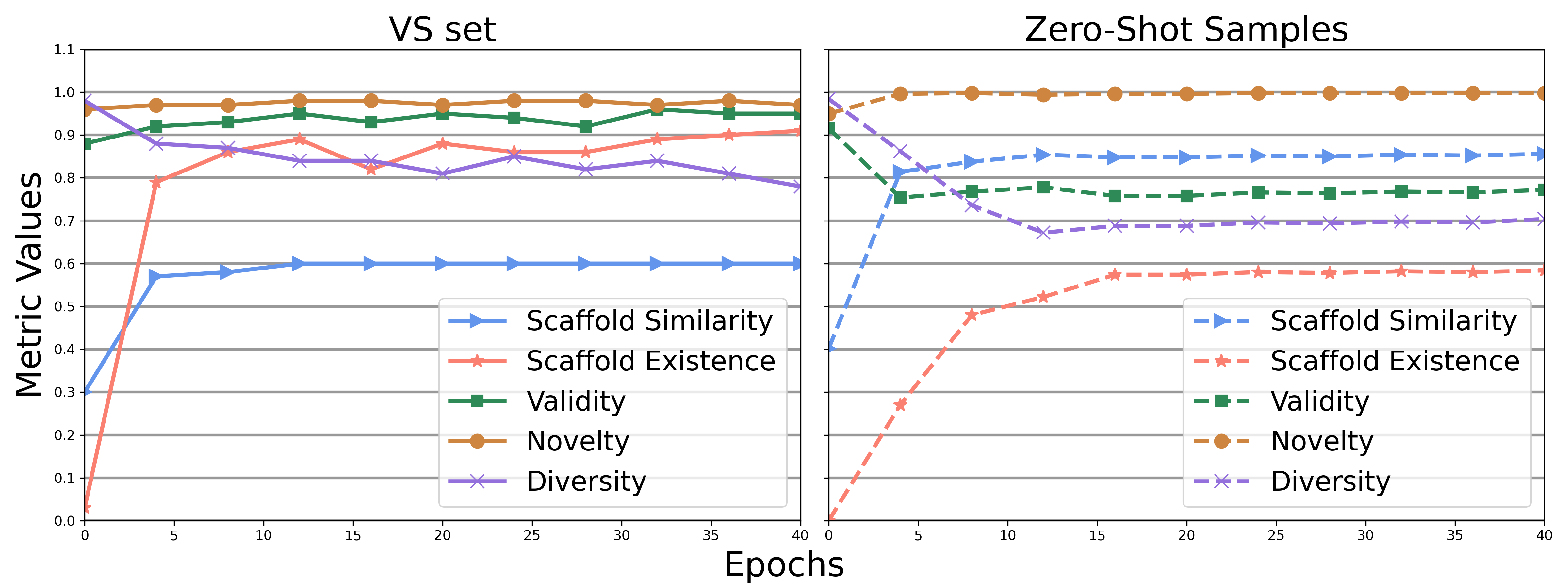}
    \caption{Evolution of SCM performance during training. The left plot shows results on the VS set, while the right plot shows results on five zero-shot scaffolds.}
    \label{fig:scm_training}
\end{figure}

Fig.~\ref{fig:scm_training} further illustrates the evolution of structural alignment during SCM training. Scaffold existence steadily improves on the VS set, while zero-shot scaffold similarity increases throughout training, demonstrating the effectiveness and generalization of SCM.

\section{Conclusion}
We propose CMCM-DLM, a flexible framework for controllable molecular generation under heterogeneous structural and property constraints. By keeping the pretrained molecular DLM frozen and introducing lightweight SCM and PCM modules, CMCM-DLM enables new control requirements to be incorporated without repeatedly retraining the full generative model, making it well suited to evolving molecular data and control scenarios. Through a two-phase diffusion process, SCM first establishes the desired molecular structure, while both SCM and PCM are subsequently combined in a common score space for joint cross-modality control. Experiments across GuacaMol, ZINC, and QM9 demonstrate that CMCM-DLM achieves strong property optimization while maintaining high scaffold adherence and generalizes across molecular datasets with different distributions and property spaces. These results highlight CMCM-DLM as an efficient and adaptable approach for controlled molecular generation over heterogeneous scientific data.

\end{document}